\documentclass{article} %
\usepackage{iclr2023_conference,times}

\usepackage{amsmath,amsfonts,bm}

\def\eqref#1{equation~\ref{#1}}

\def\1{\bm{1}}

\DeclareMathAlphabet{\mathsfit}{\encodingdefault}{\sfdefault}{m}{sl}
\SetMathAlphabet{\mathsfit}{bold}{\encodingdefault}{\sfdefault}{bx}{n}

\usepackage{hyperref}
\usepackage{url}
\usepackage{todonotes}
\usepackage{booktabs}
\usepackage{multirow}
\usepackage[capitalize]{cleveref}

\title{Polygonizer: An auto-regressive building delineator}

\author{Maxim Khomiakov$^{1,2}$,  Michael Riis Andersen$^{1}$ \& Jes Frellsen$^{1}$ \\
$^{1}$ Department of Applied Mathematics and Computer Science, Technical University of Denmark \\
$^{2}$ Otovo AS\\
\texttt{\{maxk,miri,jefr\}@dtu.dk}\\
}

\iclrfinalcopy %
\begin{document}

\maketitle
\begin{abstract}
In geospatial planning, it is often essential to represent objects in a vectorized format, as this format easily translates to downstream tasks such as web development, graphics, or design. While these problems are frequently addressed using semantic segmentation, which requires additional post-processing to vectorize objects in a non-trivial way, we present an Image-to-Sequence model that allows for direct shape inference and is ready for vector-based workflows out of the box. We demonstrate the model's performance in various ways, including perturbations to the image input that correspond to variations or artifacts commonly encountered in remote sensing applications. Our model outperforms prior works when using ground truth bounding boxes (one object per image) achieving the lowest maximum tangent angle error.
\end{abstract}

\section{Introduction}
The application of deep learning in the surveying and analysis of objects has experienced considerable advancements. Alongside the progress of general computer vision methods in classification and object detection, rapid strides have been made in the task of building delineation, which involves accurately separating building objects from the background in remote sensing imagery. However, a persisting challenge concerns the learning of realistic geometric shapes of buildings. A prevalent and intuitive initial approach is to classify the pixels associated with the object of interest using semantic segmentation. The advantages of this method include the dense, fine-grained representation achieved through pixel-by-pixel classification, while its limitations encompass high computational cost and, more significantly, the necessity for non-trivial post-processing of the predicted object masks \citep{Zorzi2020,Bittner2018}. Post-processing is essential since semantic segmentation often displays the highest uncertainty around object edges. Addressing such pixels naively with a technique like convex-hull could result in distorted objects, as even minor softening of a right-angled corner may introduce substantial artifacts upon conversion. Consequently, our work is driven by the desire to solve the building footprint delineation problem using a direct learned approach, without any post-processing.

\subsection{Related works}
A significant body of research has been published on this topic, with the majority of studies employing semantic segmentation as a fundamental component of their models. The existing literature can be classified into three categories: The first category comprises traditional computer vision approaches that utilize geometric priors and optimization algorithms to detect and polygonize buildings \citep{Li2020,Bauchet2018}. The second category encompasses studies that employ deep learning with semantic segmentation, combined with either heuristic or learned post-processing techniques \citep{Lin2015,Alidoost2019,Girard2020,Yuan2016,Bittner2018,Zorzi2020,Zhao2020,zorzi2022polyworld}. For instance, Frame Field Learning \citep{Girard2020} aims to learn effective regularization of object boundaries for precise edge definition, while PolyWorld \citep{zorzi2022polyworld} addresses the problem using a GNN model, parameterized through an adjacency matrix consisting of building corners. The third category involves research that seek to model polygons directly, thereby eliminating the necessity for post-processing \citep{Acuna2018,Li2019,Zhao2021}. Although each approach has its limitations, we argue that the benefits of minimal to no assumptions in post-processing render it the most promising direction for future research.

\section{Methods}

\begin{figure}
    \centering
    \includegraphics[width=1\textwidth]{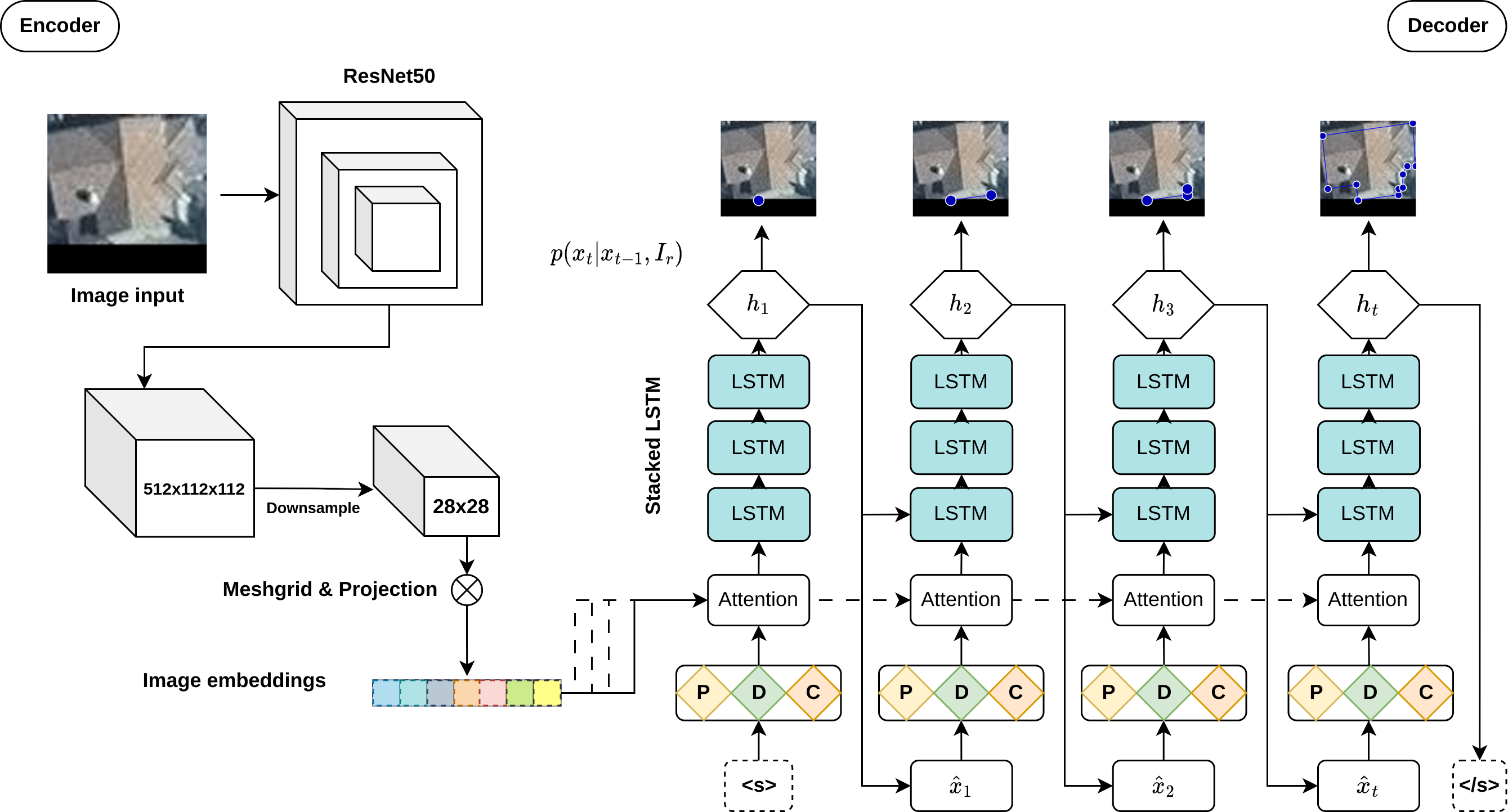}
    \caption{The encoder consists of a modified ResNet50, where the final layers are skipped, upsampled, and concatenated. Positional encoding is added to the resulting feature map, which is then fed into the decoder at each time step. The decoder processes the token at time step t, incorporating position, dimension, and coordinate embeddings along with special tokens. These special tokens include a starting token, denoted as \texttt{<s>}, and a stopping token, denoted as \texttt{</s>}.}
    \label{fig:main_figure}
\end{figure}

We present our proposed model in Figure \ref{fig:main_figure} above. Polygonizer consists of an Encoder and a Decoder, with the Encoder being a modified ResNet50, designed similarly to \citet{Acuna2018,hu2022polybuilding}, but incorporating learned embeddings for the coordinate token, coordinate dimension, and coordinate position. In addition, we introduce a small fixed value to every feature map in the 512x28x28 output generated by the encoder, which assists the network in learning spatial dependencies, resulting in our encoded feature representation $I_r$. The encoded image is then supplied as initial input, alongside a special starting token \texttt{<s>}, to a stacked LSTM Decoder, which learns to output parameters to a categorical distribution spanning the dimensions of the image $p(x_t | x_{t-1},I_r)$. At each timestep, the LSTM employs Bahdanau attention \cite{bahdanau2014neural} between the encoder representation $I_r$ and the prior hidden state of the Decoder $c_{t-1}$, aiming to align the LSTM representation with the image representation at every time step. The Decoder is trained with teacher forcing and performs inference until the special \texttt{</s>} token is encountered. While bearing similarities to the model by \citet{Li2019,Acuna2018}, our model relies solely on the preceding token and an image embedding to learn the sequence, and utilizes an LSTM instead of the ConvLSTM employed by \citet{Li2019,Acuna2018}.

\subsection{Experimental setup}
Our model is trained on the Aicrowd mapping challenge dataset \citep{mohanty2020deep} using ground truth bounding boxes. All cropped objects were padded on the smallest edge to maintain an identical aspect ratio. Our embeddings, as well as the encoder and decoder hidden dimensions, are set to 512, while the decoder is a 3-layer stacked LSTM. We optimize our model using the Adam optimizer with a negative log likelihood loss and a learning rate of $2\cdot10^{-4}$. The results shown in this paper were all obtained using greedy inference and an identical seed.

\clearpage

\section{Results}
\begin{figure}[!h]
    \centering
    \includegraphics[width=\textwidth]{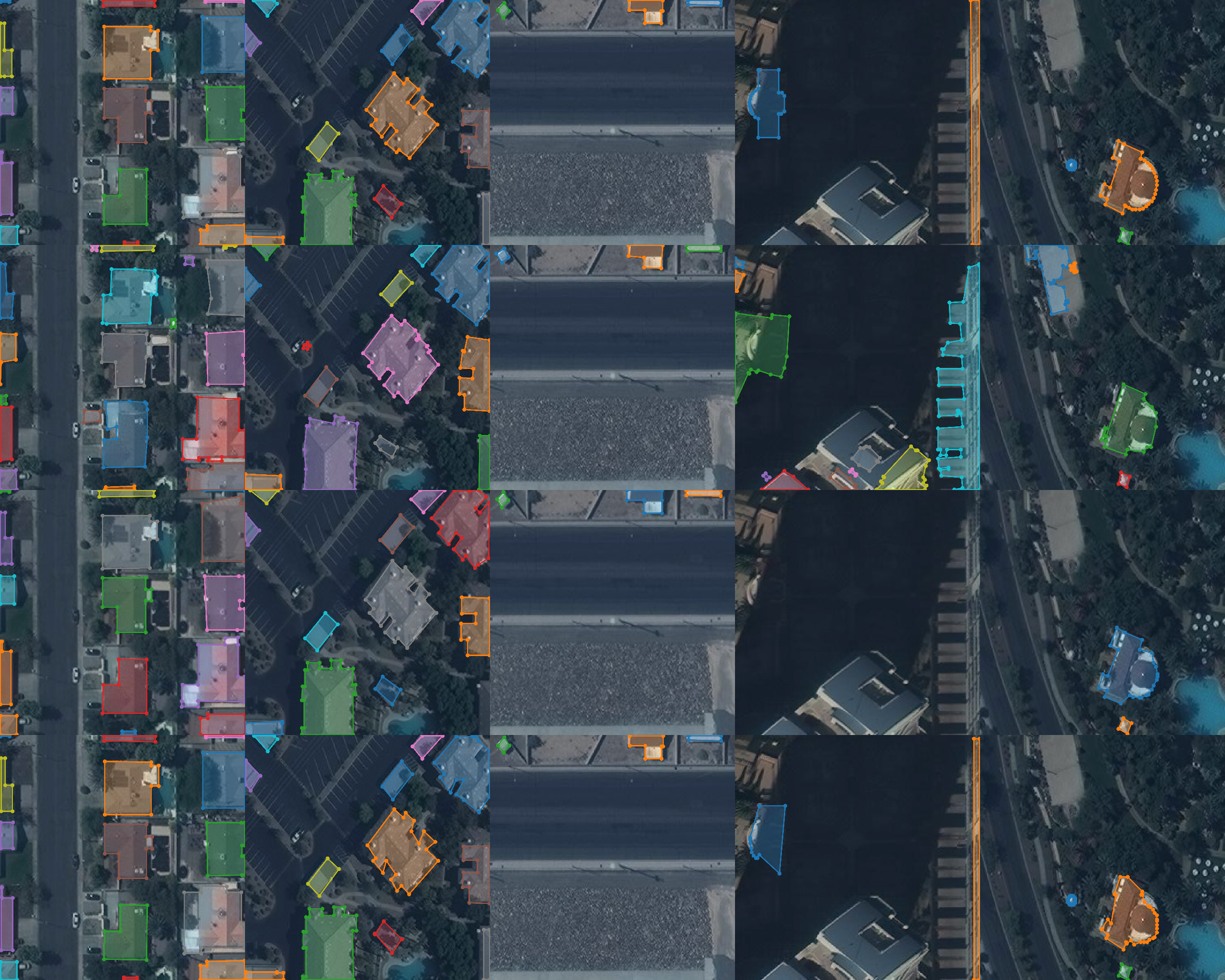}
    \caption{Qualitative comparison of results. From the top to the bottom: Ground truth, Frame Field Learning, PolyWorld and Polygonizer (ours).}
    \label{fig:results}
\end{figure}

Upon completing our research, we became aware of concurrent work by \citet{hu2022polybuilding}. As we lacked access to their model weights at the time of writing, we compared our approach with two other recent state-of-the-art methods: PolyWorld and Frame-Field Learning \citep{zorzi2022polyworld,Girard2020}. We present the COCO-evaluation metrics for an aggregate performance overview in Table \ref{fig:results}. Although our model performs reasonably well, it was trained on ground truth bounding boxes, simplifying the learning task. However, metrics like the maximum tangent angle error, one of our model's key advantages, should be less affected by this. The results in Figure \ref{fig:results} and Table \ref{tab:pertubation_performance_total} show general alignment between the methods, except for the image examples in the fourth and fifth columns. Notably, our method generally outperforms FFL and PolyWorld when using ground truth bounding boxes. However, considering the study by \cite{hu2022polybuilding}, Polygonizer ranks second or third in most metrics while demonstrating a substantial improvement over the alternative methods in having the lowest maximum tangent angle error.

\label{others}
\begin{table}[h]
\resizebox{\textwidth}{!}{\begin{tabular}{ccccccccccc}
\toprule  \toprule \text { \textbf{Method} } & {\textbf{AP}} $\uparrow$ & {\textbf{AP}} 50 $\uparrow$ & {\textbf{AP}} 75 $\uparrow$ & {\textbf{AR}} $\uparrow$ & {\textbf{AR}} 50 $\uparrow$ & {\textbf{AR}} 75 $\uparrow$ & \text { \textbf{IoU} } $\uparrow$ & {\textbf{MTA}} $\downarrow$ & {\textbf{N}} \text { \textbf{ratio} } & \text { \textbf{C-IoU} } $\uparrow$ \\
\hline \text { PolyMapper \citep{Li2019} } & 55.7 & 86 & 65.1 & 62.1 & 88.6 & 71.4 & - & - & - & - \\
\text { FFL \citep{Girard2020} } & 60.9 & 87.4 & 70.4 & 64.5 & 89.2 & 73.4 & 84.4 & 33.5 & 1.13 & 74 \\
\text { PolyWorld \citep{zorzi2022polyworld} } & 63.3 & 88.6 & 70.5 & 75.4 & 93.5 & 83.1 & 91.3 & 32.9 & 0.93 & 88.2 \\
\text { W. Li et al. \citep{li2021joint} } & 73.8 & 92 & 81.9 & 72.6 & 90.5 & 80.7 & - & - & - & - \\
 \text { PolyBuilding \citep{hu2022polybuilding} } & \textbf{78.7} & \textbf{96.3} & \textbf{89.2} & \textbf{84.2} & \textbf{97.3} & \textbf{92.9} & \textbf{94.0} & 32.4 & 0.99 & 88.6 \\ \hline
 Polygonizer (ours) & 71.9 &    94.2 &   81.4 &  82.3 &  92.9 &  90.9 &  89.8 & \textbf{10.50} & 1 & 87.9 \\
\bottomrule
\end{tabular}
}
\caption{Our model employs ground truth bounding boxes, which introduces an element of disparity in the comparison.}
\label{tab:results}
\end{table}

\subsection{A study of robustness}
Although our arguments supporting the benefits of our approach may not be definitive, we aim to investigate the extent to which comparable models can yield satisfactory results under adversarial conditions. We have devised three perturbations of the input image, which we consider to be realistic sources of error in the remote sensing domain. We evaluate the models under conditions of masking pixels (Erased dropout \citealt{zhong2020random}), downsampling the input image to simulate lower ground sampling distance, and rotating the input image in 15-degree increments. The results from these experiments are presented in Table \ref{tab:pertubation_performance_total} and in Appendix \ref{app:pertube}.
Our method demonstrates increased robustness towards downsampling and, to a lesser extent, dropout phenomena, although its performance, as expected, declines as the downsampling becomes more severe. Interestingly, PolyWorld overestimates the number of points relatively early. Both methods exhibit a significant drop in performance as the downsampling of the input image increases.

\begin{table}[htbp]
    \centering
    \begin{minipage}{0.5\textwidth}
        \centering
        \renewcommand{\arraystretch}{1.2} %
       \resizebox{0.85\textwidth}{!}{\begin{tabular}{c|rrrrrc}
    \toprule
    \toprule
    \multicolumn{1}{l|}{\multirow{6}{*}{\rotatebox[origin=c]{90}{ \textbf{PolyWorld}}}} & \textbf{AP} $\uparrow$ & \textbf{AR} $\uparrow$ & \textbf{IoU} $\uparrow$ & \textbf{N} ratio & \textbf{C-IoU} $\uparrow$ & \textbf{PF} \\ \cline{2-7}  
    \multicolumn{1}{l|}{} & 0.539 & 0.671 & 0.783 & 1.680 & 0.715 & 2 \\ 
    \multicolumn{1}{l|}{} & 0.106 & 0.256 & 0.498 & 1.143 & 0.344 & 4 \\ 
    \multicolumn{1}{l|}{} & 0.003 & 0.012 & 0.081 & 0.311 & 0.013 & 8 \\ \\ \hline
    \multicolumn{1}{l|}{\multirow{6}{*}{\rotatebox[origin=c]{90}{\textbf{FFL}}}} & \textbf{AP} $\uparrow$ & \textbf{AR} $\uparrow$ & \textbf{IoU} $\uparrow$ & \textbf{N} ratio & \textbf{C-IoU} $\uparrow$ & \textbf{PF} \\ \cline{2-7} 
    \multicolumn{1}{l|}{} & 0.531 & 0.640 & 0.781 & 0.890 & 0.713 & 2 \\ 
    \multicolumn{1}{l|}{} & 0.186 & 0.327 & 0.599 & 0.952 & 0.489 & 4 \\ 
    \multicolumn{1}{l|}{} & 0.003 & 0.030 & 0.288 & 0.552 & 0.137 & 8 \\ \\  \hline
    \multicolumn{1}{l|}{\multirow{6}{*}{\rotatebox[origin=c]{90}{\textbf{LSTM}}}} & \textbf{AP} $\uparrow$ & \textbf{AR} $\uparrow$ & \textbf{IoU} $\uparrow$ & \textbf{N} ratio & \textbf{C-IoU} $\uparrow$ & \textbf{PF} \\ \cline{2-7} 
    \multicolumn{1}{l|}{} & 0.683 & 0.797 & 0.880 & 0.963 & 0.860 & 2 \\ 
    \multicolumn{1}{l|}{} & 0.650 & 0.774 & 0.864 & 0.930 & 0.841 & 4 \\ 
    \multicolumn{1}{l|}{} & 0.523 & 0.686 & 0.801 & 0.877 & 0.761 & 8 \\ \\  %
    \bottomrule
    \end{tabular}
    }
        \label{tab:downsample}
    \end{minipage}%
    \begin{minipage}{0.5\textwidth}
        \centering
        \renewcommand{\arraystretch}{1.2} %
        \resizebox{0.85\textwidth}{!}{\begin{tabular}{c|rrrrrc}
    \toprule
    \toprule
    \multicolumn{1}{l|}{\multirow{6}{*}{\rotatebox[origin=c]{90}{\textbf{PolyWorld}}}} & \textbf{AP} $\uparrow$ & \textbf{AR} $\uparrow$ & \textbf{IoU} $\uparrow$ & \textbf{N} ratio & \textbf{C-IoU} $\uparrow$ & \textbf{PF} \\ \cline{2-7} 
    \multicolumn{1}{l|}{} & 0.448 & 0.592 & 0.747 & 1.657 & 0.667 & 4 \\ 
    \multicolumn{1}{l|}{} & 0.384 & 0.532 & 0.719 & 1.609 & 0.627 & 6 \\ 
    \multicolumn{1}{l|}{} & 0.291 & 0.444 & 0.664 & 1.517 & 0.553 & 8 \\ \\ \hline
    \multicolumn{1}{l|}{\multirow{6}{*}{\rotatebox[origin=c]{90}{\textbf{FFL}}}} & \textbf{AP} $\uparrow$ & \textbf{AR} $\uparrow$ & \textbf{IoU} $\uparrow$ & \textbf{N} ratio & \textbf{C-IoU} $\uparrow$ & \textbf{PF} \\ \cline{2-7} 
    \multicolumn{1}{l|}{} &0.443 & 0.565 & 0.739 &   0.966 & 0.678 &4 \\ 
    \multicolumn{1}{l|}{} &0.373 & 0.501 & 0.701 &   0.983 & 0.631 &6 \\ 
    \multicolumn{1}{l|}{} &0.295 & 0.432 & 0.665 &   0.990 & 0.584 &8 \\ \\
    \hline
    \multicolumn{1}{l|}{\multirow{6}{*}{\rotatebox[origin=c]{90}{\textbf{LSTM}}}} & \textbf{AP} $\uparrow$ & \textbf{AR} $\uparrow$ & \textbf{IoU} $\uparrow$ & \textbf{N} ratio & \textbf{C-IoU} $\uparrow$ & \textbf{PF} \\ \cline{2-7} 
    \multicolumn{1}{l|}{} & 0.579 & 0.725 & 0.844 &   1.086 & 0.816 & 4 \\ 
    \multicolumn{1}{l|}{} & 0.552 & 0.706 & 0.831 &   1.086 & 0.801 & 6 \\ 
    \multicolumn{1}{l|}{} &0.532 & 0.688 & 0.823 &   1.095 & 0.792 & 8 \\ \\
    \bottomrule
    \end{tabular}
    }
        \label{tab:dropout}
    \end{minipage}
    \caption{Performance relative to input pertubations. Left: Pertubation where a value of 2 equates a downsampling of 2x. Right: Erased dropout performance where the fraction of pixels dropped corresponds to PF$\cdot0.02\%$}
    \label{tab:pertubation_performance_total}
\end{table}

\section{Discussion and future works}
In this paper, we introduce a new auto-regressive method for building delineation that retains the performance of similar works \citep{Acuna2018} with reduced complexity. Our method does not necessitate a separate model to predict the first vertex and relies solely on the prior token to predict the subsequent one, unlike previous work \citep{Zhao2021,Li2019}. However, our model has its limitations. Primarily, it depends on having just one object per scene and is limited in its ability to learn long sequences. Nonetheless, the model is quite adaptable to learning right-angled geometry, which is a crucial property for remote sensing applications. Future research will focus on expanding our model to automatically detect objects in the image and introducing inductive biases that may further enhance the model's robustness.

\section{Concluding remarks}
We have proposed a novel and simplified method for polygonizing buildings from remote sensing imagery. This approach proves to be remarkably effective at learning the angles between vertices surrounding the building while also being accurate at completing the sequence. We believe this is partially explained by our fixed and learned embeddings, which, in conjunction with Bahdanau attention \citep{bahdanau2014neural} at every timestep, facilitate learning the sequential structure of the data.

\clearpage
\bibliography{iclr2023_conference}
\bibliographystyle{iclr2023_conference}

\appendix
\section{Appendix}

\section{Experimental pertubations}
\label{app:pertube}
Additional tables with experiments relating to rotation.

\begin{table}[!h]
    \centering
\resizebox{0.5\textwidth}{!}{\begin{tabular}{rrrrrc}
\toprule
\toprule
  {\textbf{AP}} $\uparrow$ &   {\textbf{AR}} $\uparrow$ &   \text { \textbf{IoU} } $\uparrow$ &  {\textbf{N}} \text { \textbf{ratio} } &  \text { \textbf{C-IoU} } $\uparrow$ &  \textbf{Rotation} \\
\midrule
\midrule
0.000 & 0.000 & 0.121 &   1.795 & 0.113 &                  90 \\
0.000 & 0.000 & 0.085 &   0.861 & 0.052 &                  60 \\
0.000 & 0.000 & 0.072 &   0.865 & 0.043 &                 120 \\
0.000 & 0.001 & 0.095 &   0.821 & 0.056 &                  45 \\
0.000 & 0.008 & 0.246 &   1.047 & 0.172 &                  15 \\
\bottomrule
\end{tabular}}
    \caption{PolyWorld \citep{zorzi2022polyworld} performance as a function of rotation.}
    \label{tab:polyworld_down}
\end{table}

\begin{table}[!h]
    \centering
\resizebox{0.5\textwidth}{!}{\begin{tabular}{rrrrrc}
\toprule
\toprule
  {\textbf{AP}} $\uparrow$ &   {\textbf{AR}} $\uparrow$ &   \text { \textbf{IoU} } $\uparrow$ &  {\textbf{N}} \text { \textbf{ratio} } &  \text { \textbf{C-IoU} } $\uparrow$ &  \textbf{Rotation} \\
\midrule
\midrule
0.003 & 0.034 & 0.336 &   0.831 & 0.311 &                  15 \\
0.000 & 0.005 & 0.136 &   0.829 & 0.122 &                  60 \\
0.000 & 0.002 & 0.115 &   0.825 & 0.103 &                 120 \\
0.000 & 0.003 & 0.132 &   0.893 & 0.123 &                  90 \\
0.000 & 0.006 & 0.151 &   0.820 & 0.134 &                  45 \\
\bottomrule
\end{tabular}}
    \caption{Frame Field Learning \cite{Girard2020} performance as a function of rotation.}
    \label{tab:polyworld-down}
\end{table}

\begin{table}[!h]
    \centering
\resizebox{0.5\textwidth}{!}{\begin{tabular}{rrrrrc}
\toprule
\toprule
  {\textbf{AP}} $\uparrow$ &   {\textbf{AR}} $\uparrow$ &   \text { \textbf{IoU} } $\uparrow$ &  {\textbf{N}} \text { \textbf{ratio} } &  \text { \textbf{C-IoU} } $\uparrow$ &  \textbf{Rotation} \\
\midrule
\midrule
0.371 & 0.510 & 0.762 &   1.252 & 0.726 &                15 \\
0.193 & 0.357 & 0.656 &   1.247 & 0.597 &                45 \\
0.167 & 0.325 & 0.637 &   1.260 & 0.578 &                60 \\
0.164 & 0.324 & 0.632 &   1.253 & 0.580 &                90 \\
0.158 & 0.319 & 0.625 &   1.259 & 0.570 &               120 \\
\bottomrule
\end{tabular}}
    \caption{Polygonizer (ours) performance as a function of rotation.}
    \label{tab:polygon}
\end{table}

\clearpage

\subsection{Additional results}

\begin{figure}[!h]
    \centering
    \includegraphics[width=\textwidth]{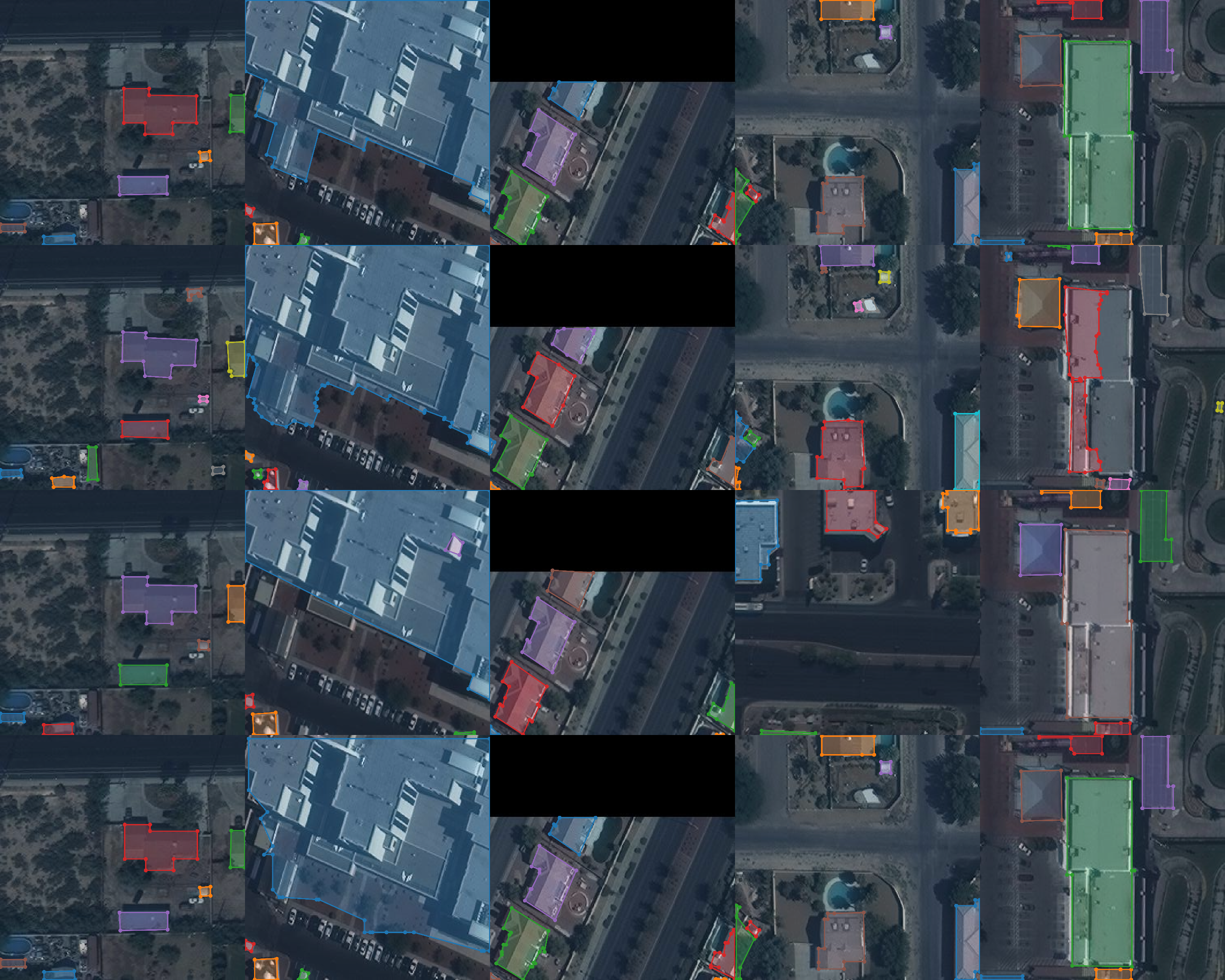}
    \caption{Qualitative comparison of results. From the top to the bottom: Ground truth, Frame Field Learning, PolyWorld and Polygonizer (ours).}
    \label{fig:appen_ex2}
\end{figure}

\begin{figure}[!h]
    \centering
    \includegraphics[width=\textwidth]{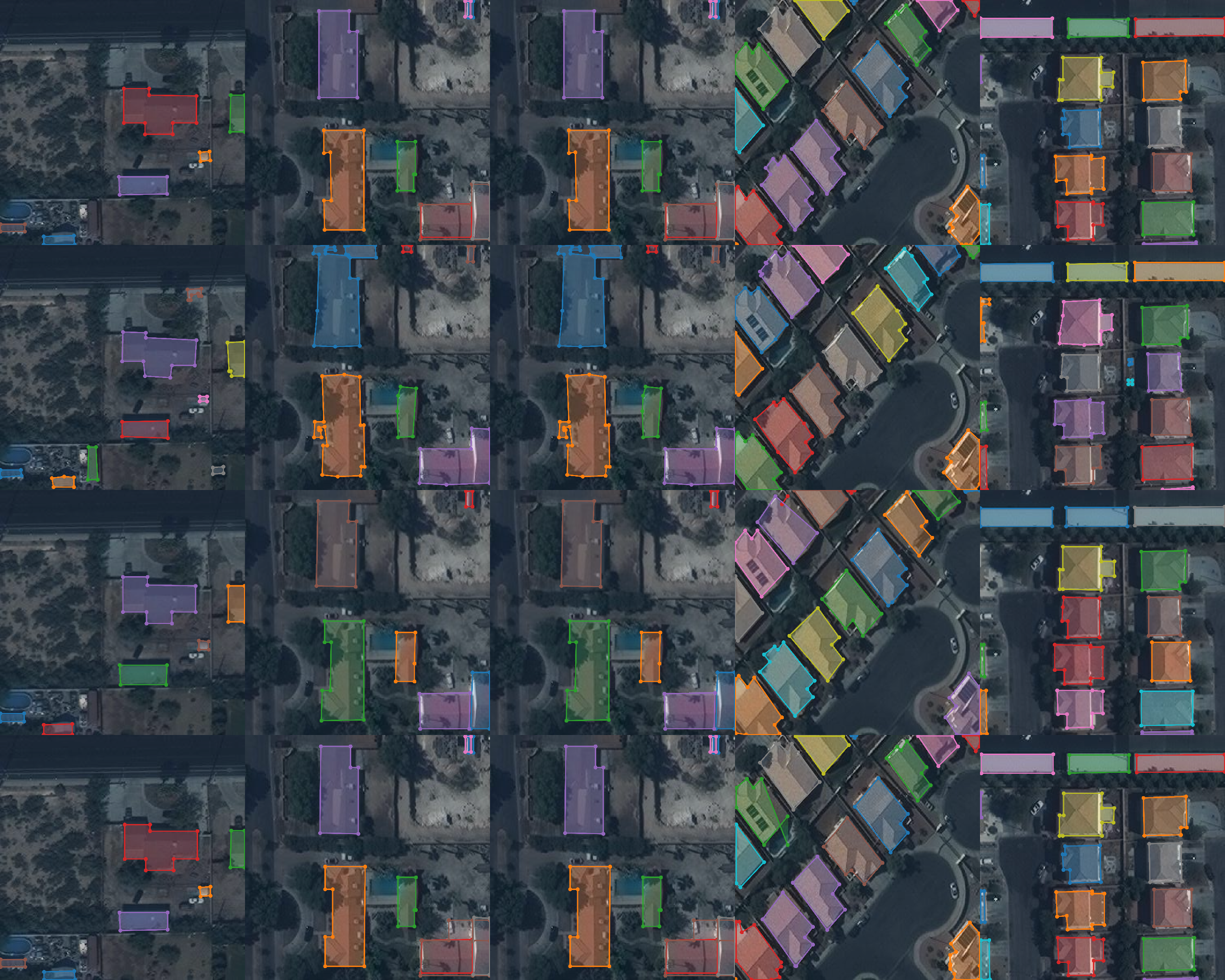}
    \caption{Qualitative comparison of results. From the top to the bottom: Ground truth, Frame Field Learning, PolyWorld and Polygonizer (ours).}
    \label{fig:appen_ex2}
\end{figure}

\begin{figure}[!h]
    \centering
    \includegraphics[width=\textwidth]{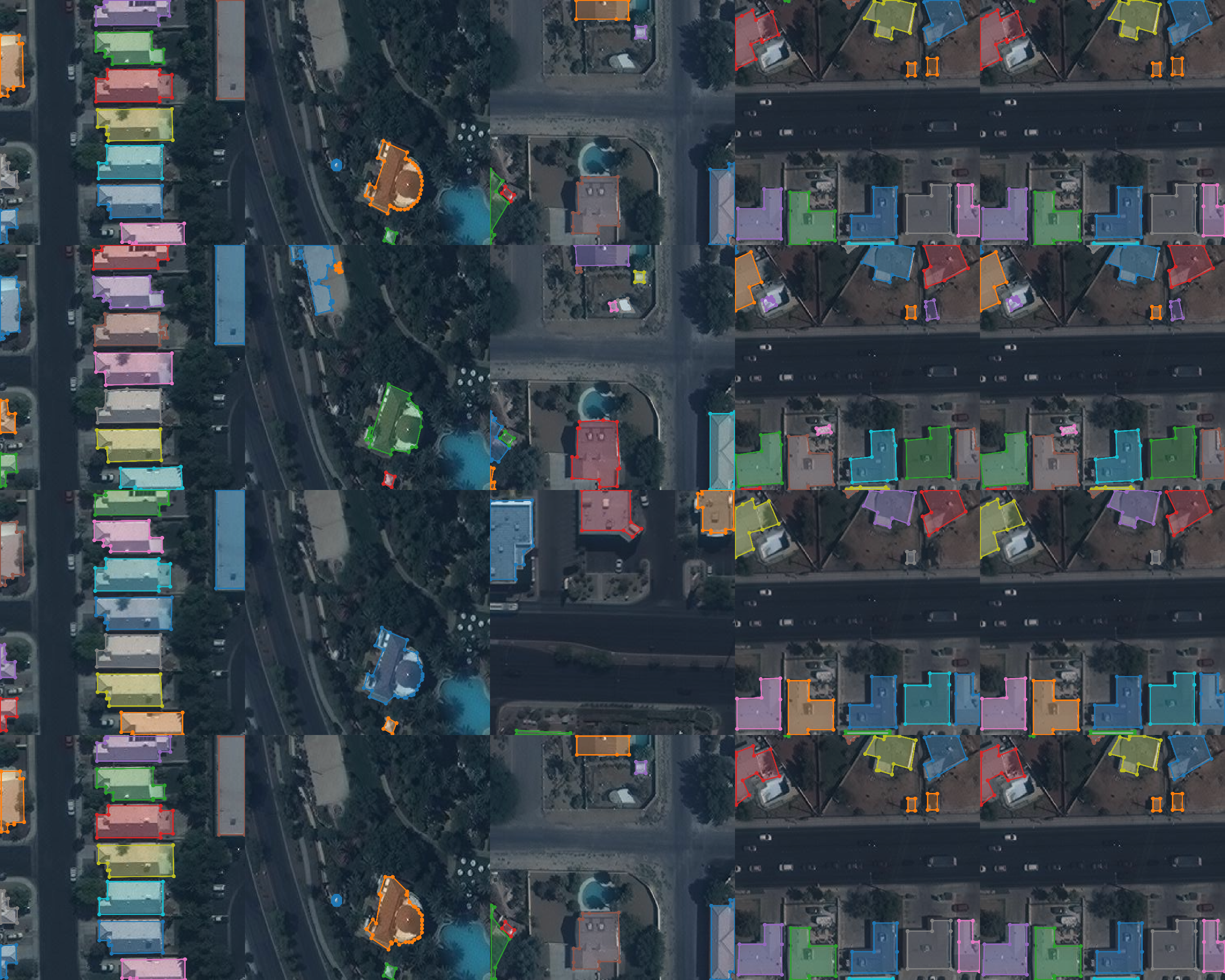}
    \caption{Qualitative comparison of results. From the top to the bottom: Ground truth, Frame Field Learning, PolyWorld and Polygonizer (ours).}
    \label{fig:appen_ex2}
\end{figure}

\end{document}